\documentclass[twocolumn,letterpaper]{IEEEAerospaceCLS}

\usepackage{amsmath}
\usepackage{graphicx}
\usepackage{url}
\usepackage{comment}
\usepackage{bm}
\usepackage{amssymb}
\usepackage{amsfonts}
\usepackage{hyperref}
\usepackage{multirow}
\usepackage{tabularx}

\newcommand{\ignore}[1]{}

\title{Architecting Autonomy for Safe Microgravity\\ Free-Flyer Inspection}

\author{
    Keenan Albee and David C. Sternberg\\
    Jet Propulsion Laboratory\\
    California Institute of Technology\\
    Pasadena, CA, USA\\
    \{keenan.albee, david.c.sternberg\}@jpl.nasa.gov
    \and
    Alexander Hansson and David Schwartz\\
    Department of Mechanical Engineering\\
    ETH Z{\"u}rich\\
    Z{\"u}rich, Switzerland\\
    \{ahansson, dschwartz\}@ethz.ch
    \vspace{1em}\and
    Ritwik Majumdar and Oliver Jia-Richards\\
    Department of Aerospace Engineering\\
    University of Michigan\\
    Ann Arbor, MI, USA\\
    \{ritwikm, oliverjr\}@umich.edu
    \thanks{\footnotesize © 2025 IEEE. Published in the Proceedings of the 2025 IEEE Aerospace Conference. Final version available at \href{https://doi.org/10.1109/AERO63441.2025.11068557}{IEEE Xplore}.}
}

\begin{document}

\maketitle
\thispagestyle{plain}
\pagestyle{plain}

\begin{abstract}
    Small free-flying spacecraft have the potential to provide vital extravehicular activity (EVA) services like inspection and repair for future orbital outposts such as the planned Lunar Gateway. Operating adjacent to delicate space station and other microgravity targets, these spacecraft require formalization to describe the autonomy that a free-flyer inspection mission must provide. This work explores the transformation of general mission requirements for this class of free-flyer into a set of concrete decisions for the planning and control autonomy architectures that will power such missions. Flowing down from operator commands for inspection of important regions and mission time-criticality, a motion planning problem emerges that provides the basis for developing autonomy solutions. Unique constraints are considered such as typical velocity limitations, pointing, and keep-in/keep-out zones, accompanied by a discussion of mission fallback techniques for providing hierarchical safety guarantees under model uncertainties and failure. Planning considerations such as cost function design and path versus trajectory control are discussed. The typical inputs and outputs of the planning and control autonomy stack of such a mission are also provided. Finally, notional system requirements such as solve times and propellant use are documented to inform planning and control design. The entire proposed autonomy framework for free-flyer inspection is realized in the SmallSatSim simulation environment, providing a reference example of free-flyer inspection autonomy. The proposed autonomy architecture serves as a blueprint for future implementations of small satellite autonomous inspection in proximity to mission-critical hardware, going beyond the existing literature in terms of both (1) providing realistic system requirements for an autonomous inspection mission and; (2) translating these requirements into autonomy design decisions for inspection planning and control.
\end{abstract} 

\vspace{-1em}
\tableofcontents

    \begin{figure}[htbp]
        \centering
        \includegraphics[width=0.45\textwidth]{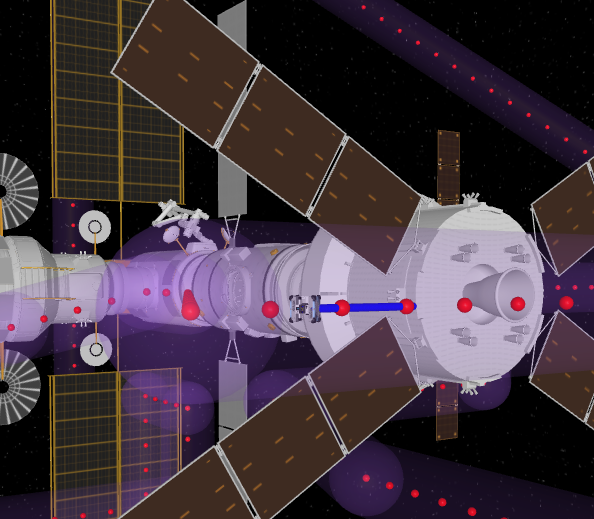} 
        \caption{Free-flyer inspection can provide agile, efficient operations in proximity to orbital outposts but will likely require autonomy for maximum safety and optimality. Here, an autonomous inspection framework is depicted in the SmallSatSim simulation environment; red circles indicate waypoints interpolated from manually-specified inspection points, $\boldsymbol{x}_{ins}$ within a purple safety corridor of radius $r_{ins}$.}
        \label{fig:main}
    \end{figure}

\section{Introduction}
\label{sec:introduction}

    
    Future orbital outposts will require regular maintenance to maintain operability. Today, external maintenance tasks on the International Space Station (ISS) are accomplished by a combination of (1) time- and risk-intensive crewed extra-vehicular activity; (2) opportunistic surveying by resupply vessels; and (3) robotic arm-aided servicing, such as the Space Station Remote Manipulator System (SSRMS) and the Special Purpose Dexterous Manipulator (DEXTRE). A key component of station maintenance is inspection to determine root causes of on-orbit anomalies. Currently, all of the above methods are used for inspection, and a combination of (1) and (3) are used for hands-on maintenance. However, these activities are accomplished only with significant human-in-the-loop (HITL) intervention, and are often limited in the agility with which they can accomplish these tasks.
    
    The proposed Lunar Gateway and other future orbital outputs will be uncrewed for long periods of time \cite{GatewayCapabilities}, inviting the use of autonomous inspection and maintenance solutions. Further, autonomous inspection will enable potentially more efficient activities (in terms of time and fuel usage) than what might be accomplished by a human teleoperator or passive imager. In lieu of other inspection modalities, free-flyer systems are well-suited to maximize field of view coverage and be quickly deployed to make time-sensitive observations. These systems will also need to operate in close proximity to their targets to see occluded regions, physically interact with payloads, and dock.
    
    With significant ground-in-the-loop time delays and agile operation in proximity to station collision volumes, autonomous free-flyer systems will need to be designed with safety as a priority, especially when considering worst-case scenarios of system failures. While prior work has studied the verification and validation philosophies that differ between small and large spacecraft \cite{GNC_VV_Philosphies_2019}, increased demands for proximity operations will require that such efforts lead to increased real-time decision making by the onboard autonomy. What's more, maintenance tasks may involve manipulating uncertain payloads that these agents must be robust to. While the benefits of these systems is clear, autonomous inspection and on-orbit handling are not well-defined; care needs to be taken in considering safety factors, easy configuration for a high-level ground operator for different inspection scenarios, as well as flexibility to accommodate different planning and control approaches and environment geometries. This work explores this autonomy architecting problem for autonomous free-flyer inspection. In creating a general planning and control framework for free-flyer inspection, this work contributes:
    
    \begin{itemize}
        \item A framework for safe, user-configurable autonomous inspection.
        \item Discussion of anticipated requirements for inspection scenarios, and their translation into the autonomy framework (e.g., time-criticality, propellant conservatism, collision avoidance, etc.).
        \item An example model predictive control approach utilizing the framework for a representative inspection scenario of Lunar Gateway.
    \end{itemize}
    
      Section \ref{sec:lit} overviews the literature on the autonomous inspection problem and on-orbit inspection in general, Section \ref{sec:mission} describes the proposed mission scenario, Section \ref{sec:approach} outlines the proposed inspection autonomy architecture, Section \ref{sec:results} provides an example of the reference case study using the architecture, and Section \ref{sec:conclusion} provides avenues for future studies. The paper therefore presents a new and versatile autonomy architecture for future inspection missions utilizing small spacecraft.


\section{Related Work}
\label{sec:lit}

    Close proximity operations allow spacecraft to operate near each others' collision volumes, a necessary requirement for a range of tasks beyond just inspection and on-orbit servicing (e.g., assembly of in-space structures or docking of the inspector spacecraft after its mission). Relevant work largely divides between a few related mission classes: inspection, docking (cooperative and non-cooperative), on-orbit assembly, and active debris removal autonomy \cite{flores-abadReviewSpaceRobotics2014} \cite{papadopoulosRoboticManipulationCapture2021}. 

    Free-flying robotic surveyors have been proposed by a number of authors to aid in these tasks \cite{davisOrbitServicingInspection2019}, and have been deployed for intra-vehicular (IVA) use including on the free-flyers SPHERES \cite{Otero2000} \cite{Sternberg2014} and Astrobee \cite{Smith2016} operating aboard the International Space Station (ISS). Astrobee continues to operate to this day as a mobile IVA astronaut assistant, engaging in a number of inspection and hands-on maintenance tasks. Other technology demonstrations such as AERCam-Sprint were motivated by IVA cargo inspection and EVA inspection of the Space Shuttle prior to reentry \cite{fredricksonApplicationMiniAERCam2004} \cite{wagenknechtDesignDevelopmentTesting2003a}. A number of additional close proximity rendezvous missions have been undertaken \cite{davisOrbitServicingInspection2019}, including autonomous rendezvous like the Mission Extension Vehicle-1 (MEV-1). Additional missions have concluded unsuccessfully after critical hardware failures, like the Seeker free-flying inspection craft \cite{pedrotty2019,pedrottySeekerFreeflyingInspector2020}.
    
    EVA free-flyer use has been limited to the aforementioned handful of restricted demonstration missions and risk-tolerant IVA use \cite{doerrReSWARMMicrogravityFlight2024}, where collisions are anticipated and non-fatal. Current state-of-practice close proximity operations remains heavily human-in-the-loop and teleoperated, such as ISS cargo resupply ingress, which limits autonomous operations to a $200$ m keep-out sphere \cite{koonsRiskMitigationApproach2010}. Semi-autonomous ingress to ISS for instance makes use of a number of carefully orchestrated hold points before and after entering a kilometers-wide approach ellipsoid, or relies on radio-assisted devices like the Soviet-derived Kurs system \cite{s.a.matviienkoOrbitalServiceSpacecraft2021}. Alternative, non-free-flyer inspection state-of-practice is heavily reliant on robotic manipulators and crewed EVAs, utilizing tools like SSRMS \cite{McGregor2001a} and DEXTRE \cite{Coleshill2009}. These alternative approaches do not offer the same mobility benefits or require direct crewmember inspection.
    
    Agile, autonomous inspection has been limited to proposals that consider the problem in environments without station-mandated risk tolerances. Yet another class of free-flyer considers passive or minimally-actuated inspection \cite{jia-richardsAnalyticalManeuverLibrary2022} \cite{Oestreich} \cite{waltonPassiveCubeSatsRemote2019}, though these techniques are limited in reuse and safety assurances. Autonomous proposals include a number of numerical optimization-based techniques \cite{ortolanoAutonomousOptimalTrajectory2021} \cite{hyeongjunparkModelPredictiveControl2014}, though these approaches either typically focus on simple collision avoidance schemes and not complex station geometry \cite{oconnorDesignImplementationSmall2012}, or focus on alternative scenarios such as tumbling target rendezvous that have different uncertainty and safety treatments \cite{albeeAutonomousRendezvousUncertain2022} \cite{Buckner2018a}. Some autonomous approaches also carefully consider uncertainty, but are not specifically tailored for close proximity inspection \cite{Nakka2021} \cite{raganOnlineTreebasedPlanning2024}; for example, a branch of approaches consider the inspection problem from a multi-agent standoff inspection perspective, usually dominated by reasoning over the relative orbital dynamics \cite{choiResilientMultiAgentCollaborative2023}.
    
    Current inspection is largely prescripted or heavily ground-in-the-loop. While safety is introduced with numerous checkpoints and ground verification time, such approaches are inefficient and not suitable for more agile and potentially non-HITL free-flyers, particularly if time-criticality or repeated inspection is desired. While some autonomous architectures have been proposed for close proximity operations, they are often not inspection-oriented or may not have fully considered relevant system constraints; on-orbit failures over the few inspection craft that have been launched, such as Seeker, show the need for caution in designing autonomous inspection approaches. This work will introduce a proposed autonomy architecture to guide principles and design choices for future autonomous inspection, starting with a number of desiderata to guide the mission formulation of Section \ref{sec:mission}.

\section{Mission Scenario}
\label{sec:mission}

    \subsection{Problem Statement: Nominal Inspection}

    The inspection problem takes the form of a constrained trajectory optimization problem,

    \begin{equation}
    \begin{aligned}
        \underset{{\boldsymbol{u}}(t)}{\min}\ J &= \int_{t_0}^{t_f} g({\boldsymbol{x}}(t), {\boldsymbol{u}}(t))\ \text{d}t + l({\boldsymbol{x}}(t_f))\\
        \text{subject to}\\
        \dot{{\boldsymbol{x}}}(t) &= f({\boldsymbol{x}}(t), {\boldsymbol{u}}(t)), t \in [0, t_f]\\
        {\boldsymbol{x}}(t_0) &= {\boldsymbol{x}}_0\\
        {\boldsymbol{x}}(t_f) &\in \mathcal{X}_N\\
        {\boldsymbol{x}}(t) &\in {\mathcal{X}}(\bm{\theta}), t \in [0, t_f]\\
        {\boldsymbol{u}}(t) &\in {\mathcal{U}}(\bm{\theta}), t \in [0, t_f],\\
    \end{aligned}
    \end{equation}
    
    where ${\boldsymbol{x}}(t) \in {\mathbb{R}}^n$ is the state, $\boldsymbol{u}(t) \in \mathbb{R}^n$ is the input, and $\mathcal{X}$ and $\mathcal{U}$ are constraint sets on the state and input. $\bm{\theta} \in \mathbb{R}^p$ is a parameter set that may modify constraints online. $g(\boldsymbol{x}(t), \boldsymbol{u}(t))$ is a running cost associated with penalties on state and input, and $l(\boldsymbol{x}(t_f))$ is a terminal cost. $\boldsymbol{x}_0$ is an initial condition constraint on the state, and $\mathcal{X}_N$ defines a goal set constraint. The dynamics, $f(\cdot)$ are frequently discretized via numerical integration resulting in the discrete trajectory optimization problem,
    
    \begin{align}
    \begin{split}
        \underset{\boldsymbol{u}_{0:N-1}}{\min} J &= \sum_{k=0}^{N-1}g(\boldsymbol{x}_k, \boldsymbol{u}_k) + l(\boldsymbol{x}_N)\\
        \text{subject to}\\
        \boldsymbol{x}_{k+1} &= f(\boldsymbol{x}_k, \boldsymbol{u}_k), \forall k \in \{0, 1, ..., N-1\}\\
        \boldsymbol{x}_0 &\triangleq \boldsymbol{x}_0\\
        \boldsymbol{x}_{N} &\in \mathcal{X}_N\\
        \boldsymbol{x}_k &\in \mathcal{X}(\bm{\theta}), \forall k \in \{0, 1, ..., N-1\}\\
        \boldsymbol{u}_k &\in \mathcal{U}(\bm{\theta}), \forall k \in \{0, 1, ..., N-1\},\\
    \label{eq:disc_traj}
    \end{split}
    \end{align}
    
    where ${\boldsymbol{x}}_k \in {\mathbb{R}}^n$ is the state, ${\boldsymbol{u}}_k \in \mathbb{R}^n$ is the input,  and $f({\boldsymbol{x}}_k, {\boldsymbol{u}}_k)$ represents the discretized dynamics. 
        
    The cost function, $J(\cdot)$ describes a planning objective for the free-flyer motion, and the constraint sets encode desired input and state safe motion. A number of desiderata exist for the inspection problem such that inspection is safe, easily performed with different potential planning techniques, easy to understand from an operator perspective, and efficient:

    \begin{itemize}
        \item Simplicity of ground operator specification: $\mathcal{X}$ and specification of $\boldsymbol{x}_{k}$ should be intuitive and easily understood to quickly produce or modify reference trajectories.
        \item Built-in safety: Interpolation between $\boldsymbol{x}_{k}$ should provide a notion of ``safety'' when in close proximity to station geometry.
        \item Field-of-view (FOV) coverage: The $\boldsymbol{x}_{\text{ref},0:N}$ specification should allow for field-of-view coverage to be specified, to enable inspection from poses of interest. Anticipated sensors include RGB and LIDAR time-of-flight depth cameras \cite{Smith2016}.
        \item Time sensitivity versus ability to linger: The approach should be adaptable to either quick, fuel-efficient trajectories (for time-sensitive investigations like a leak; or repetitive investigations, like a regular survey); or the ability to linger and seek human-in-the-loop confirmation of FOV coverage.
        \item Extendability to greater autonomy: for instance, the ability to specify a single station module view ideally would fit in easily to an autonomy inspection approach, rather than relying on greater amounts of ground operator specification.
        \item Real-time: The approach should allow for real-time computation such that plans and control actions can be computed at the time-scales of interest: $>5\ \text{Hz}$ for low-level control to match typical movement speeds.
    \end{itemize}

    Other considerations, such as motion to aid localization or account for varying lighting conditions, might also be desirable and should be captured by the autonomy architecture as potential extension modules. While localization considerations are not considered within this work, they lend interesting pathways for future investigation discussed in Section \ref{sec:conclusion}.

    Significant constraints include station geometry that must be encoded in $\mathcal{X}_{obs}$, the set of state space that cannot be entered due to obstacle collision. Additionally, the input set $\mathcal{U}$ will typically be constrained by box constraints to account for thruster saturations,

    \begin{equation}
        \label{eq:InputConstraints}
        \boldsymbol{u}_{\text{min}} \leq \boldsymbol{u}_k \leq \boldsymbol{u}_{\text{max}},\quad i = 0, \dots, N-1.
    \end{equation}

    Similar box constraints may be applied on the velocity states $\boldsymbol{v}_{\text{CoM}}^\mathcal{B}, \boldsymbol{\omega}^{\mathcal{B}}$ to limit motion to ``safe'' limits given operator risk tolerance, where CoM is defined as the center of mass. Such limits might also be used to restrict motion to aid inspection sensing quality; for example, by reducing motion blur. Additionally, the system dynamics $f(\cdot)$ are provided by the six degree of freedom (DoF) Newton-Euler equations, given as
    
    \begin{equation}
        \begin{bmatrix}
            \dot{\boldsymbol{r}}_{\text{CoM}}^\mathcal{I} \\
            \dot{\boldsymbol{q}}_{IB} \\
            \dot{\boldsymbol{v}}_{\text{CoM}}^\mathcal{B} \\
            \dot{\boldsymbol{\omega}}^\mathcal{B}   
        \end{bmatrix}
        =
        \begin{bmatrix}
            \boldsymbol{R}^{\mathcal{I}\mathcal{B}} \, \boldsymbol{v}_{\text{CoM}}^\mathcal{B} \\
            \dfrac{1}{2} \boldsymbol{H}(\boldsymbol{q}_{IB})^\top \boldsymbol{\omega}^\mathcal{B} \\
            \dfrac{1}{m} \sum_{i=1}^{n_u} \boldsymbol{F}_i - [\boldsymbol{\omega}^\mathcal{B}]_\times \boldsymbol{v}_{\text{CoM}}^\mathcal{B} \\
            \boldsymbol{I}^{-1} \left( \sum_{i=1}^{n_u} \boldsymbol{T}_i- \boldsymbol{\omega}^\mathcal{B} \times (\boldsymbol{I} \boldsymbol{\omega}^\mathcal{B}) \right)
        \end{bmatrix},
    \end{equation}
    
    where the state vector $\boldsymbol{x}$ contains
    
    \begin{equation}
        \boldsymbol{x} = \begin{bmatrix} \boldsymbol{r}_{\text{CoM}}^\mathcal{I} \\ \boldsymbol{q}_{IB} \\ \boldsymbol{v}_{\text{CoM}}^\mathcal{B} \\ \boldsymbol{\omega}^{\mathcal{B}} \end{bmatrix},
    \end{equation}
    
    whose states are defined as follows,
    
    \begin{itemize}
        \item $\boldsymbol{r}_{\text{CoM}}^\mathcal{I}$: Position vector of the system expressed in the inertial frame $\mathcal{I}$.
        \item $\boldsymbol{q}_{IB}$: Quaternion representing the orientation of the body frame $\mathcal{B}$ relative to the inertial frame $\mathcal{I}$.
        \item $\boldsymbol{v}_{\text{CoM}}^\mathcal{B}$: Linear velocity of the system expressed in the body frame $\mathcal{B}$.
        \item $\boldsymbol{\omega}^\mathcal{B}$: Angular velocity expressed in the body frame $\mathcal{B}$.
        \item $\boldsymbol{R}^{\mathcal{I}\mathcal{B}}$: Rotation matrix from the body frame $\mathcal{B}$ to the inertial frame $\mathcal{I}$.
        \item $\boldsymbol{H}(\boldsymbol{q}_{IB})$: Quaternion kinematic matrix.
        \item $[\boldsymbol{\omega}^\mathcal{B}]_\times$: Skew-symmetric matrix of $\boldsymbol{\omega}^\mathcal{B}$.
        \item $m$: Mass of the system.
        \item $\boldsymbol{F}_i$: Force produced by the $i$-th thruster, expressed in the body frame $\mathcal{B}$, for $i = 1, \dots, n_u$.
        \item $\boldsymbol{T}_i^\mathcal{B}$: Torque produced by the $i$-th thruster, expressed in the body frame $\mathcal{B}$.
        \item $\boldsymbol{I}$: Inertia matrix of the rigid body expressed in the body frame $\mathcal{B}$.
        \item $n_u$: Number of thrusters (control inputs).
    \end{itemize}
    
    Here, $\mathcal{I}$ denotes the inertial frame attached to the space station and $\mathcal{B}$ denotes the body frame.

    \subsection{Problem Statement: Model Degradation}
    \label{sec:degrade}
    
    The system model, ${\boldsymbol{x}}_{k+1} = f_k({\boldsymbol{x}}_k, {\boldsymbol{u}}_k, {\boldsymbol{w}}_k)$, ${\boldsymbol{w}}_k \in \mathbb{W}$ might be corrupted by model degradation (e.g., from a thruster failure),
    
    \begin{align}
        f(\boldsymbol{x}_k, \boldsymbol{u}_k, \boldsymbol{w}_k) = \bar{f}(\boldsymbol{x}_k, \boldsymbol{u}_k) + \hat{f}(\boldsymbol{x}_k, \boldsymbol{u}_k, \boldsymbol{w}_k),
    \end{align}

    splitting the dynamics into a nominal portion $\bar{f}(\cdot)$ and an error residual $\hat{f}(\cdot)$ corrupted by a disturbance $\boldsymbol{w}_k$ residing in disturbance set $\mathbb{W}$. Larger error residuals relative to $\bar{f}(\cdot)$ represent more difficult model degradation that must be dealt with either through robust planning and control, or inherent free-flyer trajectories that are safe-by-design. Section \ref{sec:approach} discusses a proposed planning and control autonomy framework with implicit safety that fits well into many real-time control methods that might be suitable for future work in online adaptation to account for $\hat{f}(\cdot)$.

\section{Approach: Proposed Mission Architecture}
\label{sec:approach}
    The proposed autonomy mission architecture addresses the desiderata of Section \ref{sec:mission} while solving the underlying discretized trajectory optimization problem, equation \ref{eq:disc_traj}. The approach revolves around a simple object that captures the notion of desired field of view, collision safety, and intermediate trajectory specification: inspection points,

    \begin{equation}
        \boldsymbol{x}_{ins}\triangleq \left[ \begin{array}{cc|ccccccc} t & t_{l} & r_x & r_y & r_z & q_x & q_y & q_z & q_w\end{array} \right] ^\top.
    \end{equation}

    Inspection points primarily specify pose (position $\boldsymbol{r}$ and orientation using the quaternion $\boldsymbol{q}$) at discrete points with an optional time indicator, $t$ and linger time $t_l$. There are two main modes: linger and flyby. Linger mode uses $t$ to specify a desired arrival time ($t_f$ in equation \ref{eq:disc_traj}), and the amount of time to wait at a particular waypoint, $t_l$. Linger mode may use any typical desired cost formulation $J(\cdot)$, though the traditional quadratic weighting on state ($\boldsymbol{Q}$), and input ($\boldsymbol{R}$) error is recommended for simplicity,

    \begin{align}
        g(\boldsymbol{x}_k, \boldsymbol{u}_k) &= (\boldsymbol{x}_k - \boldsymbol{x}_{k,ref})^\top \boldsymbol{Q} (\boldsymbol{x}_k - \boldsymbol{x}_{k,ref}) + \boldsymbol{u}_k^\top \boldsymbol{R} \boldsymbol{u}_k\\
        l(\boldsymbol{x}_N) &= (\boldsymbol{x}_N - \boldsymbol{x}_{N,ref})^\top \boldsymbol{Q}_N (\boldsymbol{x}_N - \boldsymbol{x}_{N,ref}).
    \end{align}
    
    Flyby mode ignores $t$ and $t_l$, taking only pose specifications, as well as potential pose derivatives $\dot{\boldsymbol{x}}_\text{ins}$, as input to the autonomous inspection framework. These poses are used as inputs to versions of $J(\cdot)$ that emphasize either minimum time, path tracking, or some combination of the above. A case study approach in Section \ref{sec:results} implements a sophisticated model predictive contouring controller, inspired by work deriving from drone racing that balances minimum time and path progression \cite{krinnerMPCCModelPredictive2024}.

    \begin{figure*}[htbp!]
        \centering
        \includegraphics[width=\textwidth]{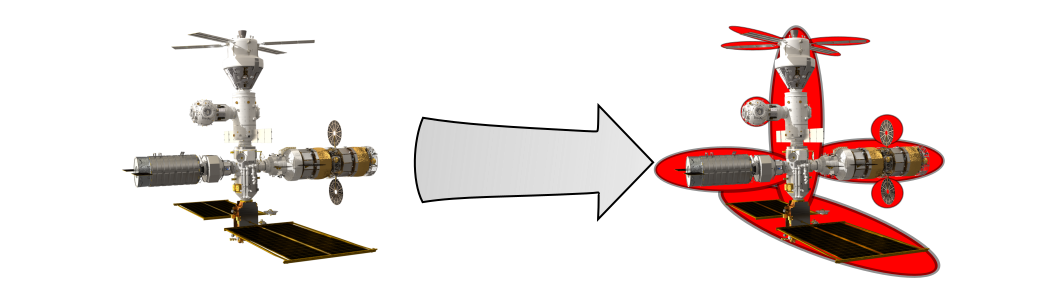} 
        \caption{$\mathcal{X}_\texttt{out}$ is efficiently constructed by taking convex hulls of major station geometry.}
        \label{fig:conv}
    \end{figure*}

    \subsection{Treatment of Obstacles in Close Proximity}

    A major challenge in moving from highly conservative circumscribing ellipsoidal station keep-out zones is efficient obstacle specification. Both linger and flyby mode proposed here rely on an efficient, simple specification of obstacle avoidance. This is accomplished by keep-in $\mathcal{X}_{in}$ and keep-out $\mathcal{X}_{out}$ zones, with $\mathcal{X}_\texttt{free} = \mathcal{X_\texttt{in} \setminus \mathcal{X}_\texttt{out}}$ defined as the set of valid state positions that are collision-free. While $\mathcal{X}_{in}$ can be specified arbitrarily, $\mathcal{X}_{out}$ is dictated by $\mathcal{X}_{obs}$, the set of inherent collision geometry defined from the station and inspector. An additional assumption is made, that the inspector robot is encased in a sphere of radius $r$, allowing simplified obstacle collision-checking calculations.


    $\mathcal{X}_{obs}$ is treated as a set of segmented convex hulls; since station geometry is well-known \textit{a priori} this segmentation can be performed manually, though automated approaches also exist for other cases involving geometries not known beforehand. Figure \ref{fig:conv} represents the convex hull segmentation approach. These convex hulls, typically ellipsoids, can be treated easily as quadratic constraints in a path planning approach \cite{jewisonModelPredictiveControl2015}. This set forms $\mathcal{X}_{out}$, which can be used in multiple potential planning approaches. The commonly used ellipsoid constraint can be represented as a quadratic constraint, useful for real-time optimization routines,

    \begin{align}
        (\boldsymbol{x} - \boldsymbol{x}_c)^\top \boldsymbol{P} (\boldsymbol{x} - \boldsymbol{x}_c) \geq 1,
    \end{align}

    where $\boldsymbol{x}_c$ is an ellipsoid centroid and $\boldsymbol{P}$ is a shaping matrix containing semi-major axis lengths.
    
\begin{figure*}[htbp!]
        \centering
        \includegraphics[width=0.5\textwidth]{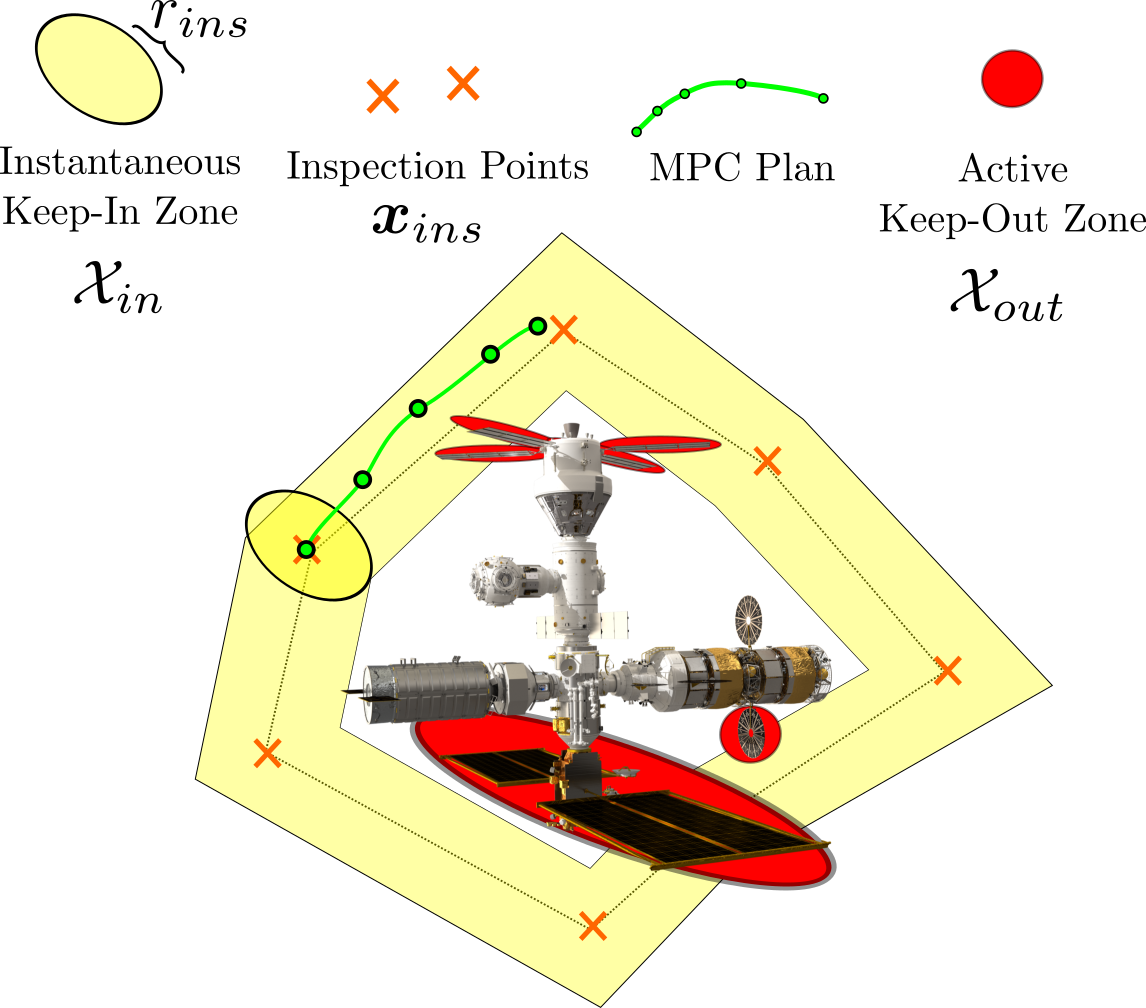} 
        \caption{The layout of the proposed autonomy framework for free-flyer inspection, consisting of inspection points that specify a desired free-flyer pose $\boldsymbol{x}_{\texttt{ins},i}$, which define keep-in corridors $\mathcal{X}_\texttt{in}$ and keep-out zones $\mathcal{X}_\texttt{out}$. The instantaneous radius $r_{ins}$ is adjusted based on the free-flyer radius $r$ and desired safe geometry clearance.}
        \label{fig:hero}
    \end{figure*}

    $\mathcal{X}_{in}$ is designed to be simple to specify, simple to use in different planning approaches, and permissive in allowing different types of inspection motion. An instantaneous radius $r_{in,k}$ is defined at each $\boldsymbol{x}_{ins,k}$, specifying a keep-in radius. These radii are linearly interpolated from $\boldsymbol{x}_{ins,k}$ to $\boldsymbol{x}_{ins,k+1}$, $\forall k \in [0, N-1]$, forming $\mathcal{X}_{in}$, shown in yellow in Figure \ref{fig:hero}. We note that inspection point specification can easily be extended to pair with a waypoint generator that performs global planning and alternative interpolation approaches, as in Figure \ref{fig:hero}. Since the proposed geometry may be non-convex, global planning could ensure plans are in the vicinity of a reasonable local minimum. $\mathcal{X}_\texttt{free}$ is then treated by the planning approach as the safe area allowed for inspection, effectively a safe tube with designated keep-out zones for closer inspection. This treatment of obstacle constraints is flexible, using a small number of basic primitives that are commonly used in planning and control approaches; it is also intuitive for ground operator interaction, requiring only specification of a desired number of $\boldsymbol{x}_{ins}$.

    Note that in the presence of model degradation, the existence of a non-zero $\hat{f}(\cdot)$ means that $\mathcal{X}_{in}$ must be carefully considered for the inaccurate \textit{a priori} model. This presents a key safety risk, as a spacecraft may not be able to enforce desired keep-in constraints under model mismatch. In practice, $\mathbb{W}$ must be bounded to make mathematical guarantees on reachability and the ability to remain within $\mathcal{X}_{in}$. Due to the unknown nature of the model degradation, these hard guarantees might become impractical; risk and collision avoidance must then be estimated and appropriate tolerances set out by mission designers in a balance against introducing excessive conservatism.

    \subsection{Tracking Inspection Points}

     There exist different ways to track the inspection points $\boldsymbol{x}_{ins}$. It is possible to track a sequence of inspection points as waypoints or to measure progress along a continuous object that passes through the waypoints. In both cases, one may also take into account the time it takes to reach certain points. If the time indicator $t$ is discarded in flyby mode, the set of waypoints forms a path, as it is a purely spatial representation and does not consider time, velocity or acceleration. However, in the case of linger mode, the set of inspection points is a time-parameterised path and thus forms a trajectory. In order to obtain a continuous trajectory, it is necessary to interpolate the waypoints, e.g. linearly, using a spline, or some other suitable parameterization. Splines offer the advantage of being smooth and maintaining higher-order continuity in the first and second derivatives (for cubic splines). This is important to have smooth velocity and acceleration profiles. However, using splines is also prone to overfitting and boundary effects. 
     An alternative to the tracking approaches mentioned above is to use reference governors \cite{Gilbert2002} \cite{Garone2017} to update the tracking point each time step. The tracking point is then set to be a a certain distance of the free-flyer along the path, and adjusted in real-time while considering the system dynamics and possible constraints. As shown in Figure \ref{fig:modules}, these approaches consider trajectory interpolation only, where $\boldsymbol{x}_{ins}$ are used directly. Alternatively, greater levels of autonomy might forego waypoint-based specification and attempt automatic trajectory \textit{generation} that attempts to automatically generate long-horizon trajectories to areas of interest while considering the FOV along the path. In all approaches, it is important to keep in mind eventual zero-velocity enforcement at the final $\boldsymbol{x}_{ins, N}$ inspection point (or, similarly, every waypoint in linger mode). The receding horizon planner must include terms for velocity enforcement at these zero-velocity points and, ideally, include multiple inspection points within a moving horizon.
     

     \subsection{To Stop or Not to Stop?}

        A key question in inspection trajectory design is whether to stop at various waypoints in linger mode or use a flyby approach. As mentioned in Section \ref{sec:mission}, this decision is driven by the amount of fuel savings relative to any safety or coverage benefits of the inspection maneuver. For instance, given inspection image exposure or LIDAR scan times, stopping might allow more thorough surveys to be conducted from a particular inspection point. A key number, then, is the typical fuel savings of a continuous approach vs. lingering approach. Disturbances that are small in magnitude with high frequency, such as reaction wheel jitter, may not pose a meaningful threat to fuel budget or the ability to provide meaningful observations \cite{ASTERIAJitter2021}, but the gross motion of the inspector requires non-negligible fuel usage. Assuming typical free-flyer mass of $10\ \text{kg}$, generous max velocity of $2\ \text{m/s}$, and representative cold-gas specific impulse of $40\ \text{s}$, the propellant required to stop and resume the trajectory per $
        \boldsymbol{x}_{ins}$ is approximately $100\ \text{g}$. Over a multi-waypoint inspection scenario, this could easily equate to $1\ \text{kg}$ or more of propellant, which will have to be weighed against the benefits of a lingering bang-off-bang approach to reaching $\boldsymbol{x}_{ins}$. It is also important to note that max inspection velocity could be further reduced as an intermediate option, requiring less fuel to stop motion, but at the cost of increased inspection time. 

    \subsection{An Autonomy Architecture for Close Proximity Inspection}

    \begin{figure*}[htbp!]
        \centering
        \includegraphics[width=0.8\textwidth]{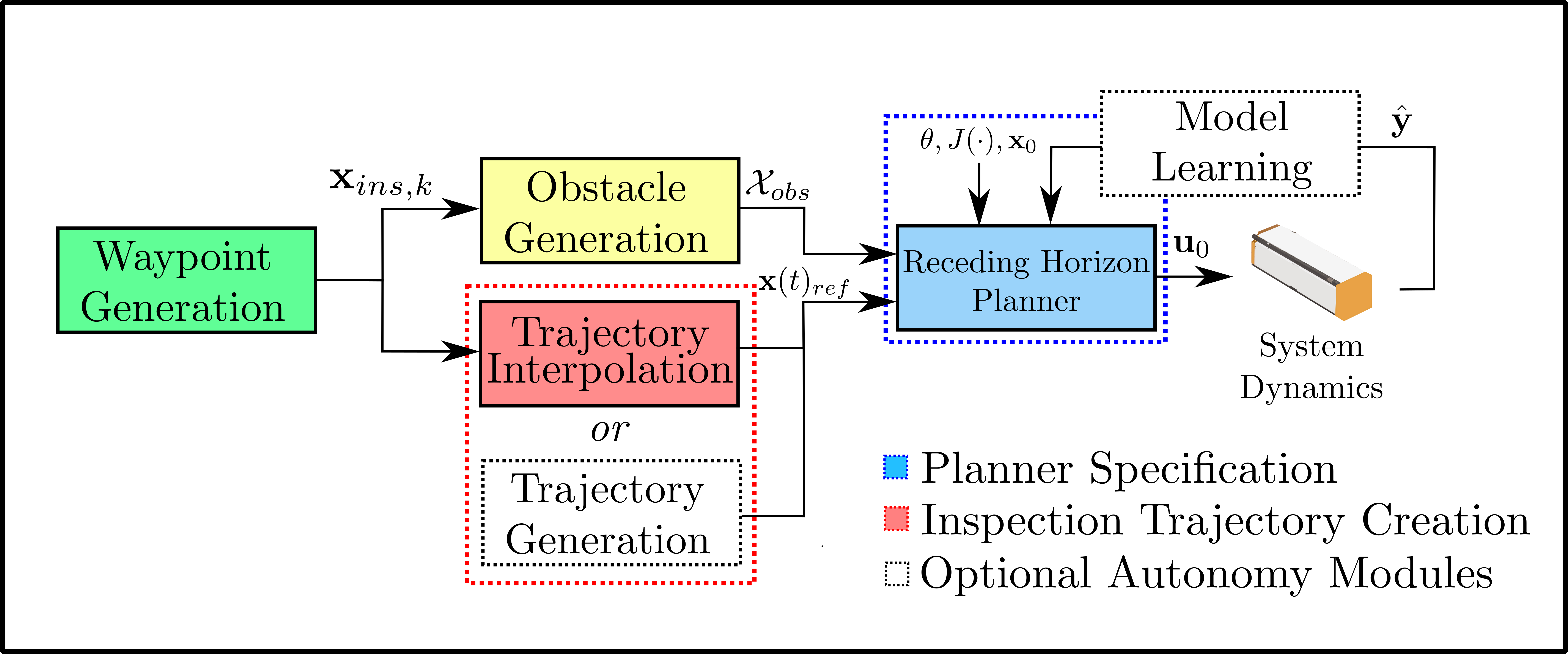} 
        \caption{The proposed autonomy framework for free-flyer inspection planning and control. Note that additional layers of autonomy, like automated trajectory generation and online adaptation, are noted as optional modules. Localization and perception are assumed to be provided by an onboard or offboard global localization system.}
        \label{fig:modules}
    \end{figure*}

    The approach of Figure \ref{fig:hero} summarizes the general composition of the receding-horizon planner $\mathcal{P}$ whose role is to provide control inputs that navigate between $\boldsymbol{x}_{ins}$. $\mathcal{X}_{free}$ is treated as described above, using the composition of keep-in and keep-out zones. Linger and flyby mode are also toggled, to determine whether a free-flyer inspector should behave using a time-efficient $J(\cdot)$. Different planning approaches may easily be substituted for $\mathcal{P}$, providing the mapping 

    \begin{align}
        \mathcal{P}(\boldsymbol{x}_{ins,k}, \mathcal{X}_{free}, \boldsymbol{x}_0, \bm{\theta}) \rightarrow \boldsymbol{u}_{0:N-1},
    \end{align}

    where $\bm{\theta}$ is the set of tuning parameters for e.g., the cost function $J(\cdot)$ to specify either linger or flyby mode. Some subset of $\boldsymbol{u}_{0:N-1}$ is executed in receding horizon fashion before recomputation of $\mathcal{P}(\cdot)$ using the latest system configuration.  The information flow of this inspection autonomy approach is specified in Figure \ref{fig:modules}, including optional modules with dotted outlines. Receding horizon planning is extremely valuable for the inspection problem, since model changes are possible as propellant is consumed, servicing targets are moved, or model changes occur. Further, availability of a localization system is assumed, providing a current and accurate state estimate $\boldsymbol{x}_0$. Note that an optional model learning module may be included, where measurements of system motion $\hat{\boldsymbol{y}}$ can be included to update planning online in the event of model changes, like the potential failures of Section \ref{sec:degrade}. Section \ref{sec:results} details an implementation of this treatment of the inspection problem using a real-time model predictive controller in linger and flyby mode.

\section{Results: An Inspection Case Study}
\label{sec:results}

    The autonomy architecture of Figures \ref{fig:hero} and \ref{fig:modules} is implemented in the MuJoCo physics engine with a sample model predictive control (MPC) scheme representing the core aspects of the autonomy framework. Flyby and linger mode are showcased, and an example of performance under severe model degradation is provided, motivating one of the primary dangers in pursuing free-flyer inspection and the subject of ongoing work. The scenarios below utilize a set of reference inspection points that might have been provided by a ground operator for a standard high coverage inspection.

    \begin{figure*}[htbp!]
        \centering
        \includegraphics[width=0.95\textwidth]{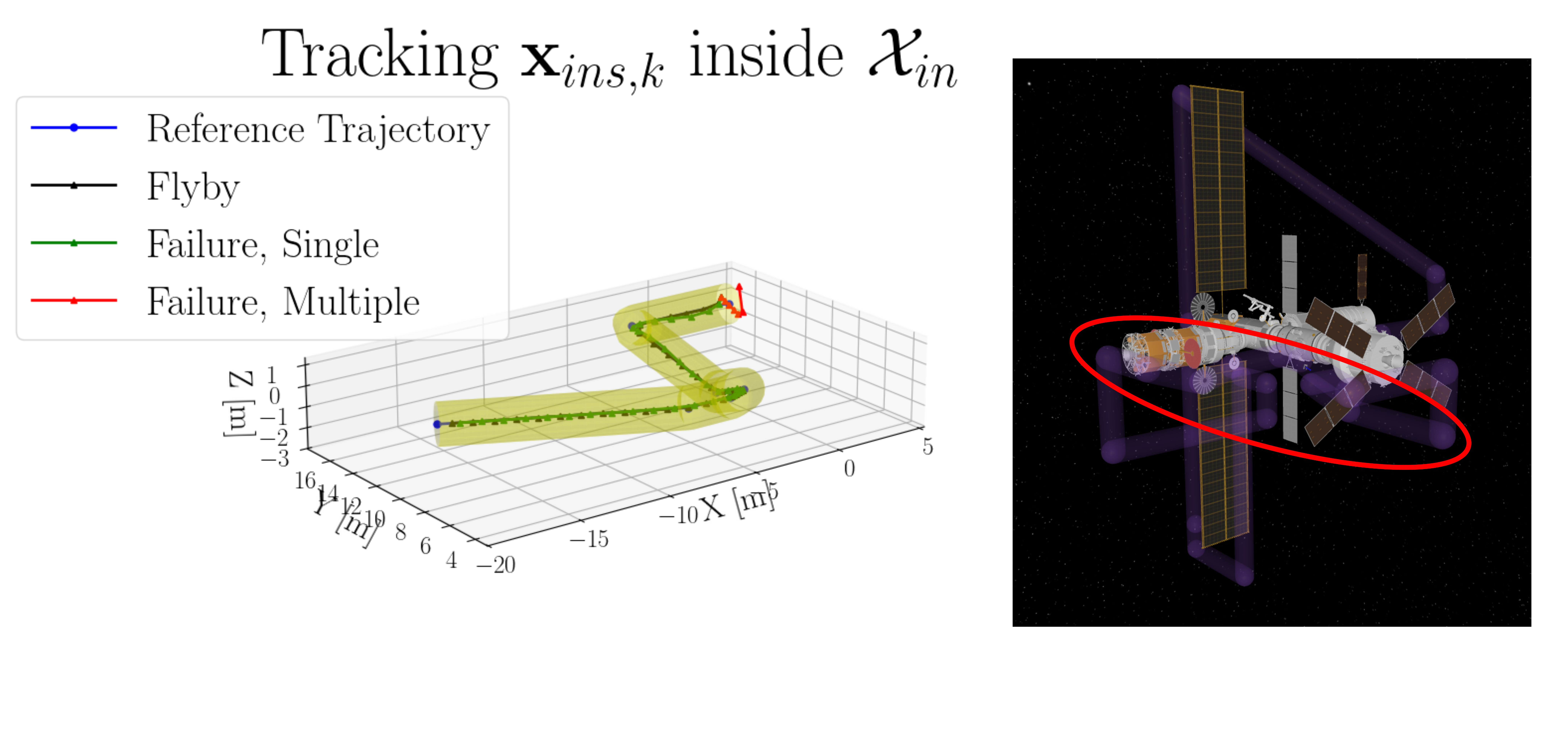} 
        \caption{Flyby trajectory tracking in the simulation environment of Figure \ref{fig:main}; the red circled portion at right is shown. $\boldsymbol{x}_{ins}$ are user-specified to provide a simple, intuitive inspection interface that can be widely integrated into inspection planning approaches. Note excellent tracking performance even with a single failure (green), and poor performance with severe model degradation (red). Note that linger mode has similar performance to flyby mode and is not shown.}
        \label{fig:flyby}
    \end{figure*}

    This reference demonstration of the inspection autonomy framework utilizes an MPC controller implemented in \texttt{acados} using a sequential quadratic programming (SQP) solver backend, achieving $>80\ \text{Hz}$ solve rates on a 3.2 GHz 8-core M1 CPU. Many implementation approaches are possible, though real-time numerical model predictive control approaches such as SQP hold particular promise in easily incorporating the desired constraint set, offering real-time speeds, and enforcing hard constraints. A model predictive contouring control (MPCC) implementation is a suitable match for flyby mode, since it penalizes continuous path progress without enforcing time constraints.

    \subsection{Inspection MPC, Flyby and Linger Mode}

    First, flyby mode is demonstrated on a 7 waypoint 41.32 m reference trajectory in proximity to approximate Lunar Gateway geometry, shown in Figure \ref{fig:flyby}. Minimum station standoff distances are made $\approx2\ \text{m}$, with $r = 0.3\ \text{m}$ and $r_{ins}$ a constant 1.0 m. A portion of the longer inspection trajectory is shown, showcasing the nominal MPC's performance in flyby mode. Under nominal model knowledge, the trajectory tracks the reference waypoints exceptionally well, with a balance between minimal thrusting and time efficiency, shown in Table \ref{tab:comparison}. Flyby mode efficiently tracks $\boldsymbol{x}_{ins}$, while maintaining trajectories within the required $\mathcal{X}_{free}$, as in Figure \ref{fig:flyby}. Linger mode is tested on the same trajectory with similar tracking performance and is not plotted for brevity. Referring again to Table \ref{tab:comparison}, linger mode's tradeoff becomes clear: total requested impulse and total time to complete the maneuver is much larger than that of flyby mode, due to the need for stops of length $t_l$ seconds at each inspection point. 
    

    \subsection{Motivating Case: Inspection MPC, under Model Degradation}

\begin{figure}[htbp!]
        \centering
        \includegraphics[width=0.45\textwidth]{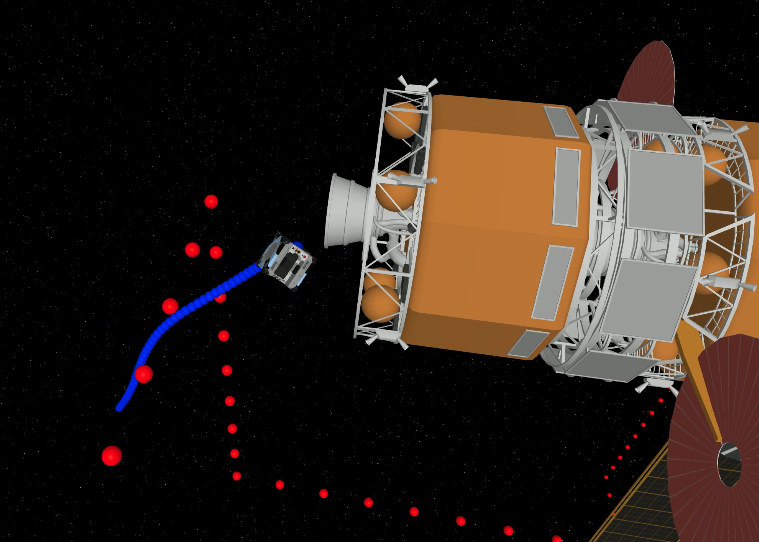} 
        \caption{Degradation of the system model via $\boldsymbol{w}_k$ can lead to dangerous behaviors without sufficient safety margin or online model learning. Here, a collision with the station geometry is shown after introducing multiple thruster failures causing the free-flyer to veer off course.}
        \label{fig:failure}
    \end{figure}

    However, a minimally conservative $\mathcal{X}_{free}$ might not be sufficient for failures or other forms of model degradation experienced during inspection. Two examples motivate this critical flaw of free-flyer inspection, and the main challenge in encouraging adoption for future use. In Figure \ref{fig:failure}, four thrusters are failed while executing the equivalent MPCC controller as above; specifically, these thrusters have all force outputs set to $0$N at $t=0$s, representing thrusters that can no longer produce output, perhaps due to a simultaneous system failure. A total of 12 thrusters placed in symmetrical force/torque pairs about the CoM are used in these experiments. Lateral deviation increases dramatically, as the nominal system model $\boldsymbol{f}(\cdot)$ fails to accurately represent the system dynamics, resulting in failure to complete the inspection trajectory and, eventually, collision shown as the red deviation in Figure \ref{fig:failure}. However, not all failures are fatal in a model-based approach: a single failure results in only mildly degraded tracking performance as shown in Figure \ref{fig:flyby} (green) and Table \ref{tab:comparison}. To account for potential catastrophic failure, waypoint generation, obstacle generation, and trajectory interpolation modules must all carefully design extra safety margin for such occurrences, or a suitable model learning module must be incorporated to allow for recovery. The key challenge in implementing safe free-flyer autonomy will remain ensuring appropriate forms of safety when operating in close proximity, and opens exciting avenues for future investigation.

    \begin{table*}[ht]
        \centering
        \caption{Comparison of Inspection Performance under Nominal and Degraded Conditions.}
        \begin{tabular}{|c|c|c|c|}
            \hline
            & \textbf{Average Lateral Dev. [m]} & \textbf{Total Translational Impulse [N-s] } & \textbf{Inspect Time [s]} \\ \hline
            \textbf{Flyby Mode, Nominal} & 0.025 &  10.29 & 165.60         \\ \hline
            \textbf{Linger Mode, Nominal} & 0.10 &  53.98 & 192.00          \\ \hline
            \textbf{Flyby Mode, 1 Failure}    & 0.065  & 18.10  & 171.50  \\ \hline
            \textbf{Flyby Mode, 4 Failure}    & 0.55  & ---  & ---     \\ \hline
        \end{tabular}
        \label{tab:comparison}
    \end{table*}

\section{Conclusion}
\label{sec:conclusion}


    Autonomy for close-proximity inspections missions is especially scrutinized for ensuring the safety of the chaser craft and the target. The risk posture for inspection missions in particular will prevent long time scales in fault detection and recovery, since the foremost goal of the inspector is to do no harm to the target. Consequently, inspection missions historically have been conducted at longer baselines---as is current state-of-practice for ISS rendezvous---with limited autonomy, or at comparatively slow velocities. The work presented in this paper discusses the considerations and constraints in an algorithmic framework to ensure such inspection autonomy can provide the necessary safety for flight implementation. Novel autonomous approaches implementing architectures similar to those proposed here will enable operation in a much more agile and responsive manner; this will enable a variety of new inspection and repair mission scenarios by decreasing relative separation distances for improved inspection sensing, increasing maximum inspector velocities for quicker inspection timelines, and improving the reliability of autonomy that can be derived from COTS or other potentially riskier hardware. The desired mission characteristics of these future inspection missions has been laid out, described in a baseline mission architecture, and demonstrated in a reference implementation. Other areas remain open for further investigation, including the perception and localization subsystems that will inform future free-flyer inspector craft in degraded lighting conditions. Future work will carefully analyze formal safety guarantees for these robotic close proximity operations, proposed methods of recovery utilizing model learning, and fully consider the case of degraded system models. 

\acknowledgments
The research was carried out at the Jet Propulsion Laboratory, California Institute of Technology, under a contract with the National Aeronautics and Space Administration (80NM0018D0004). Additional funding for this work was provided by the NASA Space Technology Mission Directorate through the University SmallSat Technology Partnerships program under grant 80NSSC23M0237. Support from the JPL Guidance and Control Spacecraft Autonomy Testbed facility personnel and Guidance and Control Section management is appreciated. The authors would also like to thank Federico Rossi and Kevin Lo for their assistance and discussion. Copyright 2025 California Institute of Technology. U.S. Government sponsorship acknowledged.

\bibliographystyle{IEEEtran}
\bibliography{references}

@inproceedings{koonsRiskMitigationApproach2010,
  title = {{Risk Mitigation Approach to Commercial Resupply to the International Space Station}},
  booktitle = {Fourth {{Annual International Association}} for the {{Advancement}} of {{Space}}},
  author = {Koons, Diane S and Schreiber, Craig and Acevedo, Francisco and Sechrist, Matt},
  year = {2010},
  month = may,
  address = {Huntsville, AL},
  langid = {english}
}

@inproceedings{krinnerMPCCModelPredictive2024,
  title = {{{{MPCC}}++: {{Model}} Predictive Contouring Control for Time-Optimal Flight with Safety Constraints}},
  shorttitle = {{{MPCC}}++},
  booktitle = {Robotics: {{Science}} and {{Systems XX}}},
  author = {Krinner, Maria and Romero, Angel and Bauersfeld, Leonard and Zeilinger, Melanie and Carron, Andrea and Scaramuzza, Davide},
  year = {2024},
  month = jul,
  publisher = {{Robotics: Science and Systems Foundation}},
  doi = {10.15607/RSS.2024.XX.109},
  urldate = {2024-10-01},
  isbn = {9798990284807},
  langid = {english}
}

@inproceedings{jewisonModelPredictiveControl2015,
  title = {{Model Predictive Control with Ellipsoid Obstacle Constraints for Spacecraft Rendezvous}},
  booktitle = {{{IFAC}}},
  author = {Jewison, Christopher and Erwin, R. Scott and {Saenz-Otero}, Alvar},
  year = {2015},
  publisher = {Elsevier Ltd.},
  doi = {10.1016/j.ifacol.2015.08.093}
}

@article{raganOnlineTreebasedPlanning2024,
  title = {{Online Tree-Based Planning for Active Spacecraft Fault Estimation and Collision Avoidance}},
  author = {Ragan, James and Riviere, Benjamin and Hadaegh, Fred Y. and Chung, Soon-Jo},
  year = {2024},
  month = aug,
  journal = {Science Robotics},
  volume = {9},
  number = {93},
  pages = {eadn4722},
  issn = {2470-9476},
  doi = {10.1126/scirobotics.adn4722},
  urldate = {2024-10-01},
  langid = {english}
}

@inproceedings{choiResilientMultiAgentCollaborative2023,
  title = {{Resilient {{Multi-Agent Collaborative Spacecraft Inspection}}}},
  booktitle = {2023 {{IEEE Aerospace Conference}}},
  author = {Choi, Changrak and Nakka, Yashwanth Kumar and Rahmani, Amir and Chung, Soon-Jo},
  year = {2023},
  month = mar,
  pages = {1--10},
  publisher = {IEEE},
  address = {Big Sky, MT, USA},
  doi = {10.1109/AERO55745.2023.10115886},
  urldate = {2024-04-25},
  copyright = {https://doi.org/10.15223/policy-029},
  isbn = {978-1-66549-032-0},
  langid = {english}
}

@article{hyeongjunparkModelPredictiveControl2014,
  title = {{Model Predictive Control for Spacecraft Rendezvous and Docking with a Rotating/Tumbling Platform and for Debris Avoidance}},
  author = {{Hyeongjun Park} and Di Cairano, Stefano and Kolmanovsky, Ilya},
  year = {2014},
  journal = {American Control Conference},
  pages = {1922--1927},
  doi = {10.1109/acc.2011.5991151},
  isbn = {9781457700811}
}

@inproceedings{s.a.matviienkoOrbitalServiceSpacecraft2021,
  title = {{Orbital Service Spacecraft}},
  booktitle = {International {{Astroanutical Congress}} ({{IAC}})},
  author = {{S.A. Matviienko}},
  year = {2021},
  month = oct,
  address = {Dubai, UAE}
}

@inproceedings{Smith2016,
  title = {{Astrobee: A New Platform for Free-Flying Robotics on the {{ISS}}}},
  booktitle = {International {{Symposium}} on {{Artificial Intelligence}}, {{Robotics}} and {{Automation}} in {{Space}} (i-{{SAIRAS}})},
  author = {Smith, Trey and Barlow, Jonathan and Bualat, Maria and Fong, Terrence and Provencher, Christopher and Sanchez, Hugo and Smith, Ernest},
  year = {2016},
  address = {Beijing, China}
}

@inproceedings{albeeAutonomousRendezvousUncertain2022,
  title = {{Autonomous Rendezvous with an Uncertain, Uncooperative Tumbling Target: The {{TumbleDock}} Flight Experiments}},
  booktitle = {{{ESA ASTRA}}},
  author = {Albee, Keenan and Specht, Caroline and Mishra, Hrishik and Oestreich, Charles and Brunner, Bernhard and Linares, Richard},
  year = {2022},
  pages = {8},
  address = {Noordwijk, The Netherlands},
  copyright = {All rights reserved},
  langid = {english}
}

@inproceedings{GNC_VV_Philosphies_2019,
  title = {{Adaptations of Guidance, Navigation and Control Verification and Validation Philosophies for Small Spacecraft}},
  booktitle = {2019 {{42nd AAS Annual Guidance and Control Conference}}},
  author = {Pong, Christopher and Sternberg, David and Chen, George},
  year = {2019},
  month = jan,
  pages = {1--13},
  publisher = {AAS},
  address = {Breckenridge, CO, USA},
  langid = {english},
}

@inproceedings{ASTERIAJitter2021,
  title = {{Validation of Small Satellite Dynamics Simulation Modules using ASTERIA Flight Data}},
  booktitle = {2021 {{IEEE Aerospace Conference}}},
  author = {Sternberg, David and Mohan, Swati},
  year = {2021},
  month = mar,
  pages = {1--14},
  publisher = {IEEE},
  address = {Big Sky, MT, USA},
  langid = {english},
}

@phdthesis{Buckner2018a,
  title = {{Tube-Based Model Predictive Control for the Approach Maneuver of a Spacecraft to a Free-Tumbling Target Satellite}},
  author = {Buckner, Caroline and Lampariello, Roberto},
  year = {2018},
  volume = {2018-June},
  issn = {07431619},
  doi = {10.23919/ACC.2018.8431558},
  isbn = {9781538654286}
}

@article{Coleshill2009,
  title = {{{{DEXTRE}}: {{Improving}} Maintenance Operations on the {{International Space Station}}}},
  author = {Coleshill, Elliott and Oshinowo, Layi and Rembala, Richard and Bina, Bardia and Rey, Daniel and Sindelar, Shelley},
  year = {2009},
  journal = {Acta Astronautica},
  volume = {64},
  number = {9-10},
  pages = {869--874},
  issn = {00945765},
  doi = {10.1016/j.actaastro.2008.11.011}
}

@misc{davisOrbitServicingInspection2019,
  title = {{On-Orbit Servicing: Inspection, Repair, Refuel, Upgrade, and Assembly of Satellites in Space}},
  author = {Davis, Joshua P and Mayberry, John P and Penn, Jay P},
  year = {2019},
  publisher = {{Aerospace Center for Space Policy and Strategy}},
  langid = {english}
}

@article{doerrReSWARMMicrogravityFlight2024,
  title = {{The {{ReSWARM}} Microgravity Flight Experiments: {{Planning}}, Control, and Model Estimation for On-orbit Close Proximity Operations}},
  shorttitle = {The {{ReSWARM}} Microgravity Flight Experiments},
  author = {Doerr, Bryce and Albee, Keenan and Ekal, Monica and Ventura, Rodrigo and Linares, Richard},
  year = {2024},
  month = apr,
  journal = {Journal of Field Robotics},
  pages = {rob.22308},
  issn = {1556-4959, 1556-4967},
  doi = {10.1002/rob.22308},
  urldate = {2024-04-22},
  copyright = {All rights reserved},
  langid = {english}
}

@article{flores-abadReviewSpaceRobotics2014,
  title = {{A Review of Space Robotics Technologies for On-Orbit Servicing}},
  author = {{Flores-Abad}, Angel and Ma, Ou and Pham, Khanh and Ulrich, Steve},
  year = {2014},
  journal = {Progress in Aerospace Sciences},
  volume = {68},
  pages = {1--26},
  issn = {03760421},
  doi = {10.1016/j.paerosci.2014.03.002},
  urldate = {2017-11-02},
  isbn = {0376-0421}
}

@inproceedings{fredricksonApplicationMiniAERCam2004,
  title = {{Application of the Mini {{AERCam}} Free Flyer for Orbital Inspection}},
  booktitle = {Defense and {{Security}}},
  author = {Fredrickson, Steven E. and Duran, Steve and Howard, Nathan and Wagenknecht, Jennifer D.},
  editor = {Tchoryk, Jr., Peter and Wright, Melissa},
  year = {2004},
  month = aug,
  pages = {26--35},
  address = {Orlando, FL},
  doi = {10.1117/12.542810},
  urldate = {2023-05-03},
  langid = {english}
}

@article{jia-richardsAnalyticalManeuverLibrary2022,
  title = {{Analytical Maneuver Library for Remote Inspection with an Underactuated Spacecraft}},
  author = {{Jia-Richards}, Oliver and Lozano, Paulo C.},
  year = {2022},
  month = apr,
  journal = {Journal of Guidance, Control, and Dynamics},
  volume = {45},
  number = {4},
  pages = {611--622},
  issn = {1533-3884},
  doi = {10.2514/1.G005766},
  urldate = {2024-04-15},
  langid = {english}
}

@article{McGregor2001a,
  title = {{Flight 6a: Deployment and Checkout of the {{Space Station Remote Manipulator System}} ({{SSRMS}})}},
  author = {McGregor, Rod and Oshinowo, Layi},
  year = {2001},
  journal = {Proceeding of the 6th International Symposium on Artificial Intelligence and Robotics \& Automation in Space: i-SAIRAS 2001},
  number = {May}
}

@phdthesis{Nakka2021,
  title = {{Spacecraft Motion Planning and Control under Probabilistic Uncertainty for Coordinated Inspection and Safe Learning}},
  author = {Nakka, Yashwanth Kumar},
  year = {2021},
  school = {California Institute of Technology}
}

@mastersthesis{oconnorDesignImplementationSmall2012,
  title = {{Design and Implementation of Small Satellite Inspection Missions}},
  author = {O'Connor, Michael C.},
    school = {Massachusetts Institute of Technology},
  year = {2012},
  volume = {66},
  number = {June},
  isbn = {9786026258076}
}

@inproceedings{Oestreich,
  title = {{On-Orbit Inspection of an Unknown, Tumbling Target Using {{NASA}}'s {{Astrobee}} Robotic Free-Flyers}},
  booktitle = {Conference on {{Computer Vision}} and {{Pattern Recognition}}},
  author = {Oestreich, Charles and Teran, Antonio and Todd, Jessica and Albee, Keenan and Linares, Richard},
  year = {2021},
  address = {Virtual},
  copyright = {All rights reserved}
}

@article{ortolanoAutonomousOptimalTrajectory2021,
  title = {{Autonomous Optimal Trajectory Planning for Orbital Rendezvous, Satellite Inspection, and Final Approach Based on Convex Optimization}},
  author = {Ortolano, Nicholas and Geller, David K. and Avery, Aaron},
  year = {2021},
  month = jun,
  journal = {The Journal of the Astronautical Sciences},
  volume = {68},
  number = {2},
  pages = {444--479},
  issn = {0021-9142, 2195-0571},
  doi = {10.1007/s40295-021-00260-5},
  urldate = {2024-09-30},
  langid = {english}
}

@phdthesis{Otero2000,
  title = {{The {{SPHERES}} Satellite Formation Flight Testbed: Design and Initial Control}},
  author = {{Saenz-Otero}, Alvar},
  year = {2000},
  urldate = {2018-01-22},
  school = {Massachusetts Institute of Technology}
}

@mastersthesis{Sternberg2014,
  title = {{Development of an Incremental and Iterative Risk Reduction Facility for Robotic Servicing and Assembly Missions}},
  author = {{Sternberg}, David},
  year = {2014},
  school = {Massachusetts Institute of Technology},
}

@article{papadopoulosRoboticManipulationCapture2021,
  title = {{Robotic Manipulation andt Capture in Space: A Survey}},
  shorttitle = {Robotic Manipulation and Capture in Space},
  author = {Papadopoulos, Evangelos and Aghili, Farhad and Ma, Ou and Lampariello, Roberto},
  year = {2021},
  month = jul,
  journal = {Frontiers in Robotics and AI},
  volume = {8},
  pages = {686723},
  issn = {2296-9144},
  doi = {10.3389/frobt.2021.686723},
  urldate = {2022-02-10},
  langid = {english}
}

@inproceedings{pedrotty2019,
    author = {Samuel Pedrotty and Jacob Sullivan and Elisabeth Gambone and Thomas Kirven},
    title = {{Seeker Free-Flying Inspector GNC System Overview}},
    booktitle = {Proceedings of the AAS Guidance and Control Conference},
    publisher = {American Astronautical Society (AAS)},
    number = {AAS 19-156},
    year = {2019},
    address = {Breckenridge, Colorado, USA},
}

@inproceedings{pedrottySeekerFreeflyingInspector2020,
    author = {Pedrotty, Samuel and Sullivan, Jacob and Gambone, Elisabeth and Kirven, Thomas},
    title = {{Seeker Free-Flying Inspector GNC Flight Performance}},
    booktitle = {Proceedings of the AAS Guidance and Control Conference},
    publisher = {American Astronautical Society (AAS)},
    number = {AAS 20-158},
    year = {2020},
    address = {Breckenridge, Colorado, USA}
}

@inproceedings{wagenknechtDesignDevelopmentTesting2003a,
  title = {{Design, Development and Testing of the {{Miniature Autonomous Extravehicular Robotic Camera}} ({{Mini AERCam}}) Guidance, Navigation, and Control System}},
  booktitle = {American {{Astronautical Society}}},
  author = {Wagenknecht, J},
  year = {2003},
  urldate = {2018-01-08}
}

@article{waltonPassiveCubeSatsRemote2019,
  title = {{Passive {{CubeSats}} for Remote Inspection of Space Vehicles}},
  author = {Walton, Patrick and Cannon, Josh and Damitz, Brian and Downs, Tyler and Glick, Dallon and Holtom, Jacob and Kohls, Nicholas and Laraway, Alex and Matheson, Iggy and Redding, Jason and Robinson, Cory and Ryan, Jared and Stoddard, Niall and Willis, Jacob and Warnick, Karl and Wirthlin, Michael and Wilde, Doran and Iverson, Brian D. and Long, David},
  year = {2019},
  month = jul,
  journal = {Journal of Applied Remote Sensing},
  volume = {13},
  number = {03},
  pages = {1},
  issn = {1931-3195},
  doi = {10.1117/1.JRS.13.032505},
  urldate = {2024-09-30},
  langid = {english}
}

@article{Gilbert2002,
   author = {Elmer Gilbert and Ilya Kolmanovsky},
   doi = {10.1016/S0005-1098(02)00135-8},
   issn = {00051098},
   issue = {12},
   journal = {Automatica},
   title = {{Nonlinear Tracking Control in the Presence of State and Control Constraints: A Generalized Reference Governor}},
   volume = {38},
   year = {2002},
}

@article{Garone2017,
   author = {Emanuele Garone and Stefano Di Cairano and Ilya Kolmanovsky},
   doi = {10.1016/j.automatica.2016.08.013},
   issn = {00051098},
   journal = {Automatica},
   title = {{Reference and Command Governors for Systems with Constraints: A Survey on Theory and Applications}},
   volume = {75},
   year = {2017},
}

@online{GatewayCapabilities,
    author = "NASA",
    title = "Snapshot: Gateway Capabilities",
    url  = "https://www.nasa.gov/gateway-capabilities/",
    addendum = "(accessed: 11.20.2024)",
}

\thebiography
\begin{biographywithpic}
    {Keenan Albee}{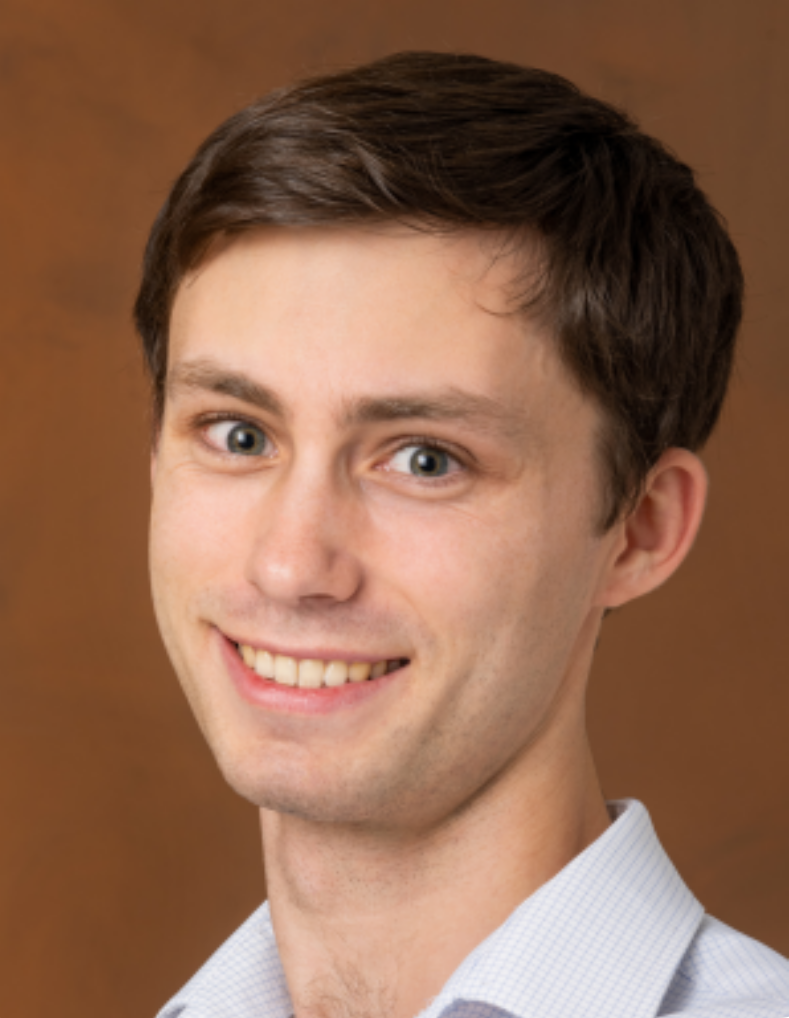} is a Robotics Technologist in the Maritime and Multi-Agent Autonomy group at NASA's Jet Propulsion Laboratory. Keenan received a Ph.D. in Aeronautics and Astronautics (Autonomous Systems) from MIT in 2022 under a NASA Space Technology Research Fellowship. Keenan's research focuses on planning and control for mobile robotic systems operating in uncertain environments, leveraging real-time tools to make autonomous robotic operations safer and more efficient.
\end{biographywithpic} 

\begin{biographywithpic}
    {David C. Sternberg}{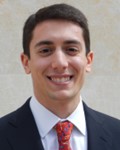} is a guidance and control systems engineer at the NASA Jet Propulsion Laboratory, having earned his SB, SM, and ScD degrees in the MIT Department of Aeronautics and Astronautics. He is currently working on the development, testing, and operation of attitude determination and control systems for several satellites, particularly the Psyche mission, as well as the creation of various spacecraft testbeds and simulations. His doctoral work in Space Systems Engineering focused on the development of optimal trajectories for docking to tumbling targets with uncertain properties.
\end{biographywithpic}

\begin{biographywithpic}
    {Alexander Hansson}{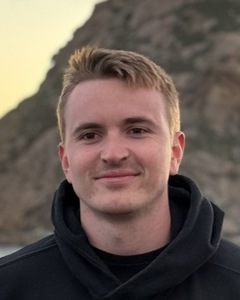} is a Master's student at ETH Zurich, studying Robotics, Systems, and Control. His research interests are in the fields of controls, machine learning and computer vision.
\end{biographywithpic}
\vspace{4em}

\begin{biographywithpic}
    {David Schwartz}{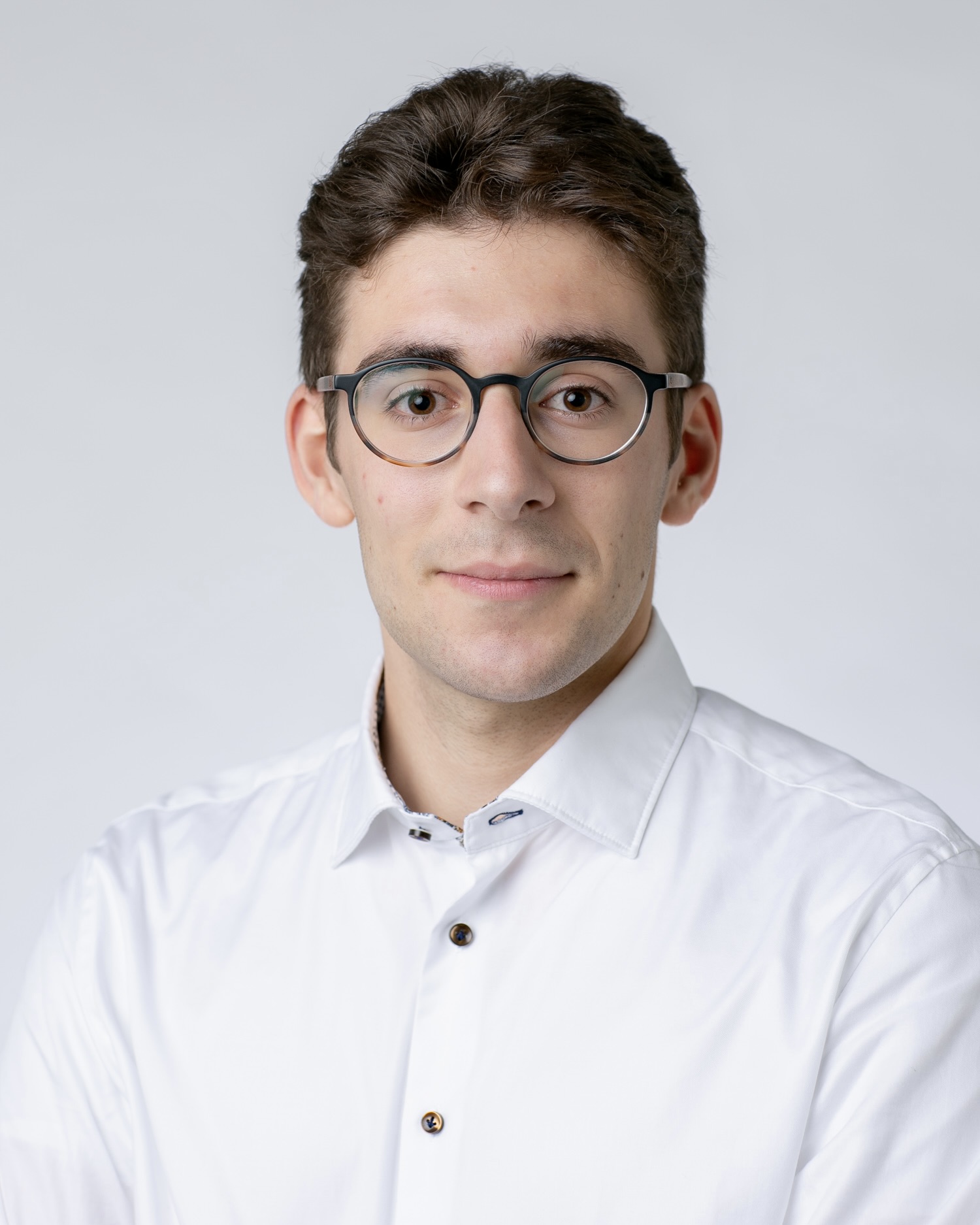} is a Master's student at ETH Zurich studying Mechanical Engineering. His research interests lie at the intersection of learning-based planning/control under uncertainty and safety.
\end{biographywithpic}
\vspace{3em}

\begin{biographywithpic}
    {Ritwik Majumdar}{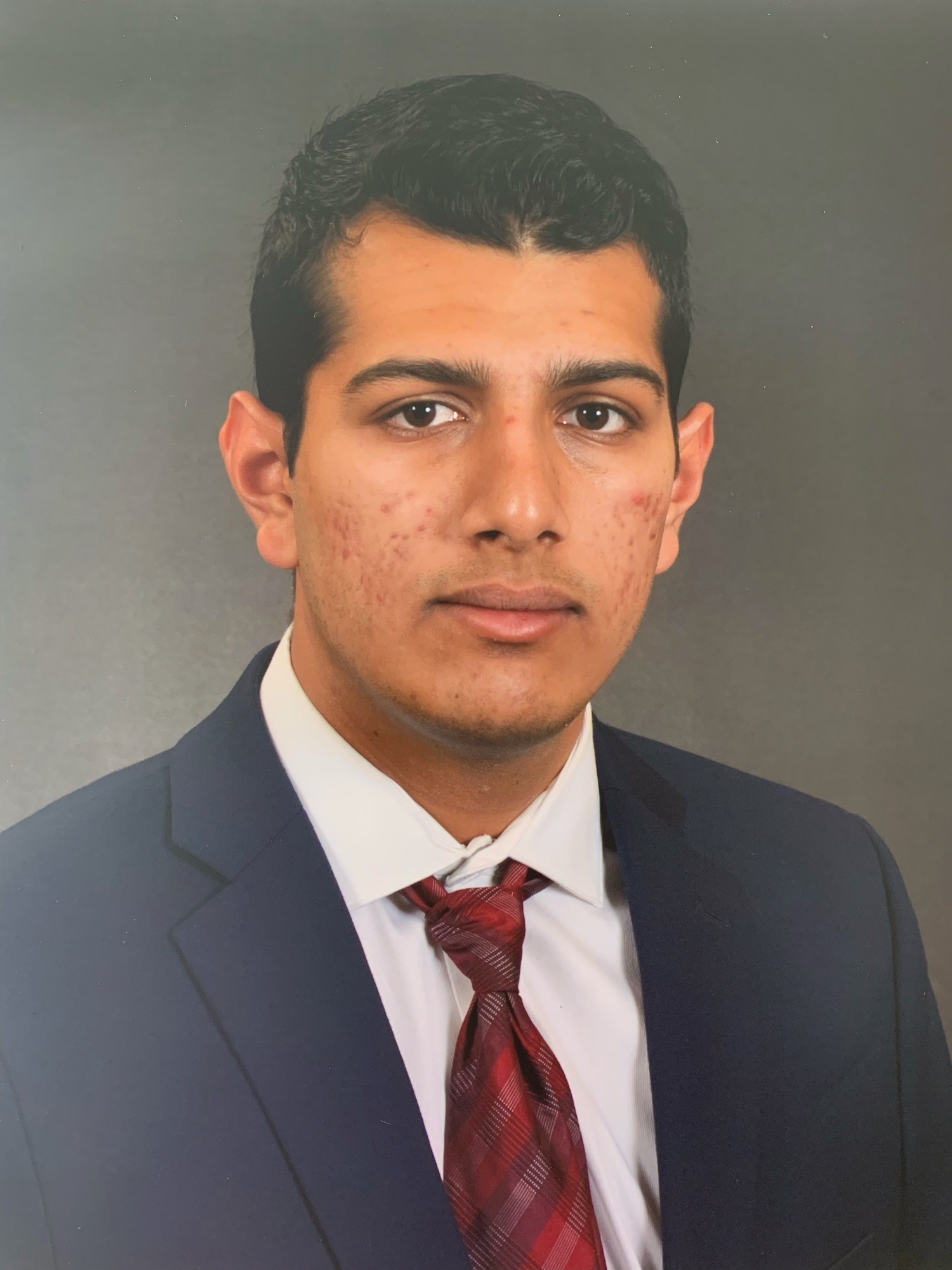} is a PhD student studying aerospace engineering at the University of Michigan. His research interests are in the field of machine learning and space systems. Specifically in the application of reinforcement learning methods to autonomous spacecraft controls.
\end{biographywithpic}
\vspace{1em}

\begin{biographywithpic}
    {Oliver Jia-Richards}{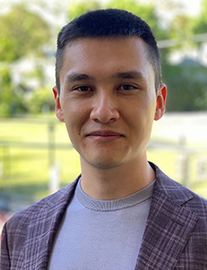} is an Assistant Professor of Aerospace Engineering at the University of Michigan. His research interests lie in the domain of space propulsion and spaceflight mechanics, with a primary interest in modeling, characterizing, and manipulating the performance of propulsion systems, particularly electric propulsion systems, in order to improve spacecraft guidance and control.
\end{biographywithpic}

\end{document}